\documentclass{bmvc2k}
\usepackage{amsmath}
\usepackage{amssymb}
\usepackage{graphicx}
\usepackage[percent]{overpic}
\usepackage{multirow}
\usepackage[justification=centerlast]{floatrow}
\usepackage{xspace}

\newfloatcommand{capbtabbox}{table}[][\FBwidth]

\newcommand{\dataset}{\emph{FaceValue}\xspace}
\newcommand{\bff}{\mathbf{f}}
\renewcommand{\paragraph}[1]{\par\smallskip\noindent{\bf #1}}

\title{Learning Grimaces by Watching TV}
\addauthor{Samuel Albanie}{http://www.robots.ox.ac.uk/~albanie}{1}
\addauthor{Andrea Vedaldi}{http://www.robots.ox.ac.uk/~vedaldi}{1}
\addinstitution{
 Engineering Science Department\\
 Univeristy of Oxford\\
 Oxford, UK
}
\runninghead{Albanie, Vedaldi}{Learning Grimaces by Watching TV}

\def\eg{\emph{e.g}\bmvaOneDot}

\def\etal{\emph{et al}\bmvaOneDot}

\begin{document}
% ---------------------------------------------------------
\maketitle
\begin{abstract}
Differently from computer vision systems which require explicit supervision, humans can learn facial expressions by observing people in their environment. In this paper, we look at how similar capabilities could be developed in machine vision. As a starting point, we consider the problem of relating facial expressions to objectively-measurable events occurring in videos. In particular, we consider a gameshow in which contestants play to win significant sums of money. We extract events affecting the game and corresponding facial expressions objectively and automatically from the videos, obtaining large quantities of labelled data for our study. We also develop, using benchmarks such as FER and SFEW 2.0, state-of-the-art deep neural networks for facial expression recognition, showing that pre-training on face verification data can be highly beneficial for this task. Then, we extend these models to use facial expressions to predict events in videos and learn nameable expressions from them.  The dataset and emotion recognition models are available at \url{http://www.robots.ox.ac.uk/~vgg/data/facevalue}.
\end{abstract}

% ---------------------------------------------------------
\section{Introduction}\label{sec:intro}
% ---------------------------------------------------------

Humans make extensive use of facial expressions in order to communicate. Facial expressions are complementary to other channels such as speech and gestures, and often convey information that cannot be recovered from the other two alone. Thus, understanding facial expressions is often necessary to properly understand images and videos of people.

The general approach to facial expression recognition is to label a dataset of faces with either \emph{nameable expressions} (\eg happiness, sadness, disgust, anger, etc.) or \emph{facial action units} (movements of facial muscles such as tightening the lips or raising an upper eyelid) and then learn a corresponding classifier, for example by using a deep neural network. In contrast, humans need not to be \emph{explicitly told} what facial expressions means, but can learn that by associating facial expressions to how people react to particular events or situations.\footnote{Generating certain facial expressions is an innate ability; however, recognizing facial expression is a learned skill.} 

\begin{figure}[t]
\begin{center}
\includegraphics[width=0.333\textwidth,clip,trim=0 42 0 42]{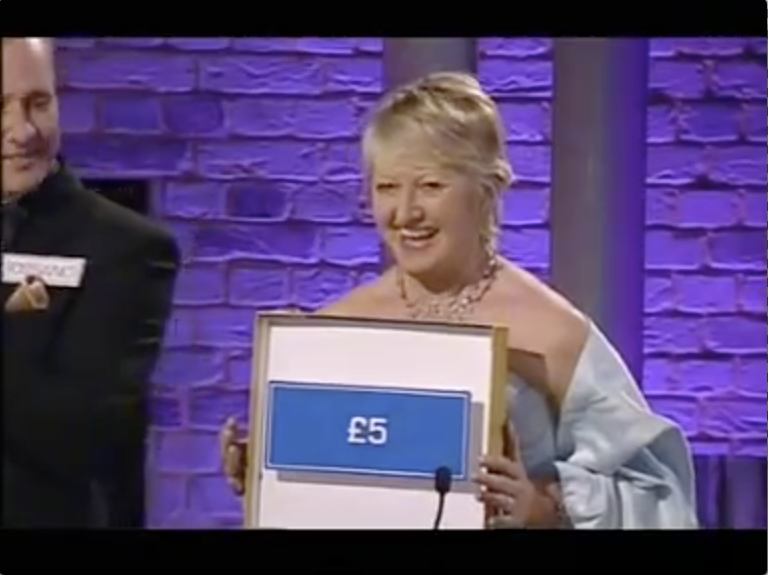}%
\includegraphics[width=0.333\textwidth,clip,trim=0 42 0 42]{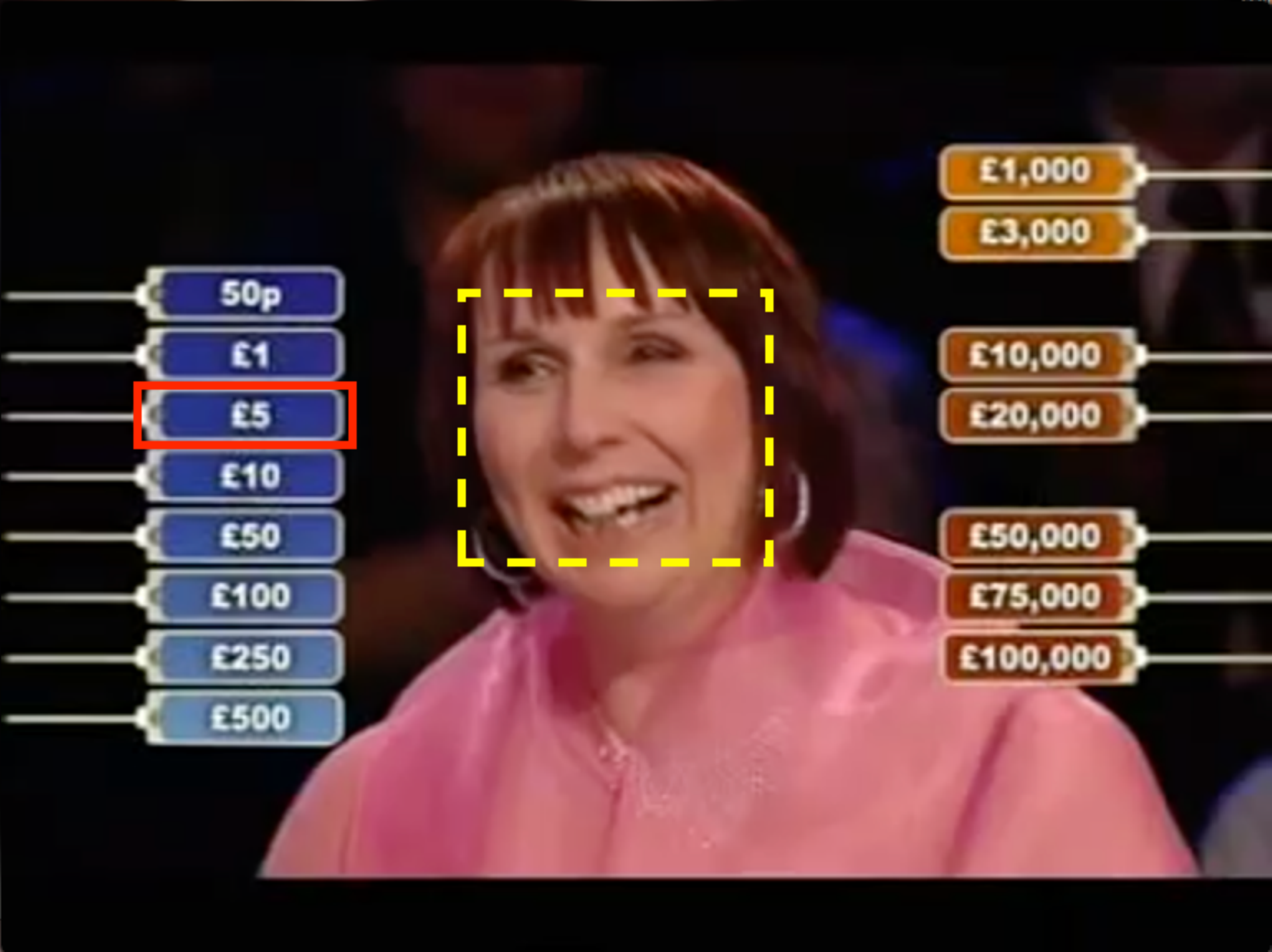}%
\includegraphics[width=0.333\textwidth,clip,trim=0 42 0 42]{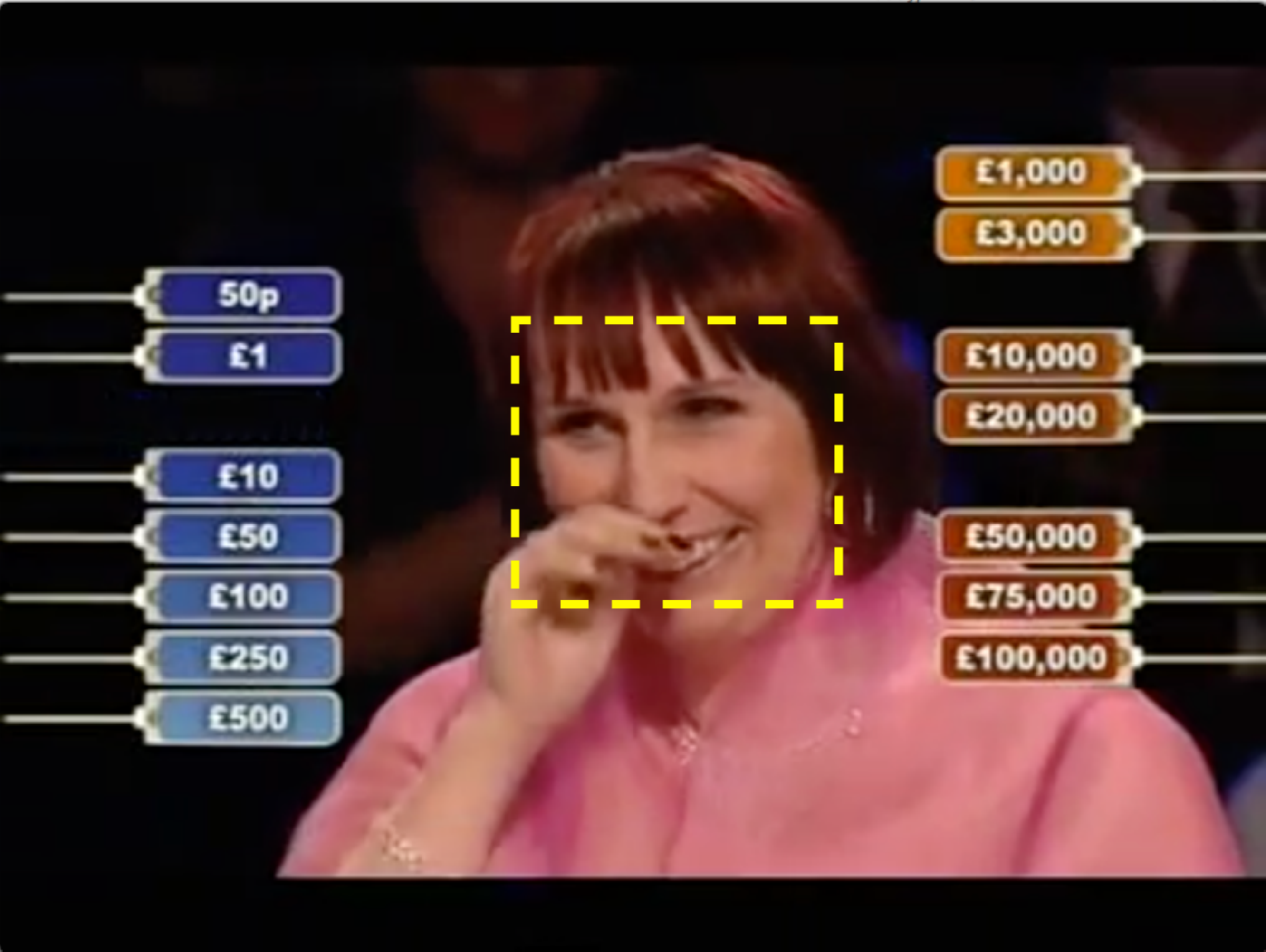}\\
\includegraphics[width=0.08333\textwidth,height=0.08333\textwidth]{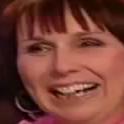}%
\includegraphics[width=0.08333\textwidth,height=0.08333\textwidth]{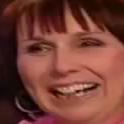}%
\includegraphics[width=0.08333\textwidth,height=0.08333\textwidth]{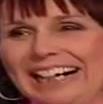}%
\includegraphics[width=0.08333\textwidth,height=0.08333\textwidth]{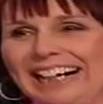}%
\includegraphics[width=0.08333\textwidth,height=0.08333\textwidth]{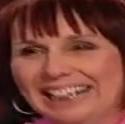}%
\includegraphics[width=0.08333\textwidth,height=0.08333\textwidth]{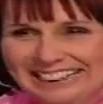}%
\includegraphics[width=0.08333\textwidth,height=0.08333\textwidth]{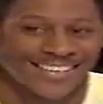}%
\includegraphics[width=0.08333\textwidth,height=0.08333\textwidth]{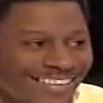}%
\includegraphics[width=0.08333\textwidth,height=0.08333\textwidth]{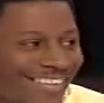}%
\includegraphics[width=0.08333\textwidth,height=0.08333\textwidth]{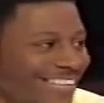}%
\includegraphics[width=0.08333\textwidth,height=0.08333\textwidth]{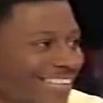}%
\includegraphics[width=0.08333\textwidth,height=0.08333\textwidth]{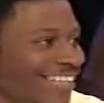}
\includegraphics[width=0.08333\textwidth,height=0.08333\textwidth]{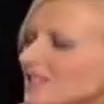}%
\includegraphics[width=0.08333\textwidth,height=0.08333\textwidth]{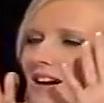}%
\includegraphics[width=0.08333\textwidth,height=0.08333\textwidth]{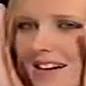}%
\includegraphics[width=0.08333\textwidth,height=0.08333\textwidth]{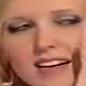}%
\includegraphics[width=0.08333\textwidth,height=0.08333\textwidth]{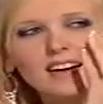}%
\includegraphics[width=0.08333\textwidth,height=0.08333\textwidth]{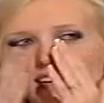}%
\includegraphics[width=0.08333\textwidth,height=0.08333\textwidth]{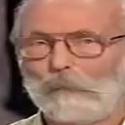}%
\includegraphics[width=0.08333\textwidth,height=0.08333\textwidth]{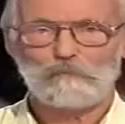}%
\includegraphics[width=0.08333\textwidth,height=0.08333\textwidth]{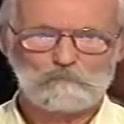}%
\includegraphics[width=0.08333\textwidth,height=0.08333\textwidth]{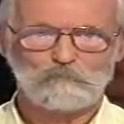}%
\includegraphics[width=0.08333\textwidth,height=0.08333\textwidth]{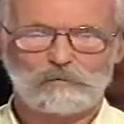}%
\includegraphics[width=0.08333\textwidth,height=0.08333\textwidth]{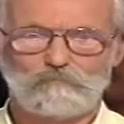}%
\end{center}
\vspace{-2em}
\caption{\textbf{\dataset} dataset. We study facial expressions from objectively-measurable events occurring in the ``Deal or No Deal'' gameshow. \emph{Top:} detection of an event at round $t=6$ in the game. Left: a box is opened, revealing to the contestant that her prize is \emph{not} the one of value $x_t=\pounds 5$. Since this is a low amount, well below the expected value of the prize of $E_5 = \pounds 17,331$, this is a ``good'' event for the contestant. Right: the contestant's face, intuitively expressing happiness, is detected. Note also the overlay for $x_t=\pounds 5$ disappearing from a frame to the next; our system can automatically read such cues to track the state of the game. \emph{Bottom:} four example tracks, the top two for ``good'' events and the bottom two for ``bad'' events, as defined in the text.}\label{f:splash}
\end{figure}

In order to investigate whether algorithms can also learn facial expressions by establishing similar associations, in this paper we look at the problem of \emph{relating facial expressions to objectively-quantifiable contextual events in videos}. The main difficulty of this task is that there is only a weak correlation between an event occurring in a video and a person showing a particular facial expression. However, learning facial expressions in this manner has three important benefits. The first one is that it grounds the problem on objectively-measurable quantities, whereas labelling emotions or even facial action units is often ambiguous. The second benefit is that contextual information can often be labelled in videos fully or partially automatically, obviating the cost of collecting large quantities of human-annotated data for data-hungry machine learning algorithms. Finally, the third advantage is that the ultimate goal of face recognition in applications is not so much to describe a face, but to infer from it information about a situation or event, which is tackled directly by our study.

Concretely, our {first contribution} (Sect.~\ref{s:data}; Fig.~\ref{f:splash}) is to develop a novel dataset, \dataset, of faces extracted from videos together with objectively-measurable contextual events. The dataset is based on the ``Deal or No Deal'' TV program, a popular game where contestants can win or lose significant sums of money. Using a semi-automatic procedure, we extract significant events in the game along with the player (and public) reaction. We use this data to predict from facial expressions whether events are ``good'' or ``bad'' for the contestant. To the best of our knowledge, this is the first example of leveraging gameshows in facial expression understanding and the first study aiming to relate facial expressions to people's activities.

Our {second contribution} is to carefully assess the difficulty of this problem by establishing a human baseline and by extending the latter to existing expression recognition datasets for comparison (Sect.~\ref{s:human}). We also develop a number of \emph{state-of-the-art expression recognition models} (Sect.~\ref{s:networks}) and show that excellent performance can be obtained by transferring deep neural networks from face verification to expression recognition. Our final contribution is to extend such systems to the problem of recognising \dataset events from facial expressions (Sect.~\ref{s:context}). We develop simple but effective pooling strategies to handle face tracks, integrating them in deep neural network architectures. With these, we show that it is not only possible to predict events from facial expressions, but also to learn nameable expressions by looking at people spontaneously reacting to events in TV programs.

% ---------------------------------------------------------
\subsection{Related work}\label{s:related}
% ---------------------------------------------------------

\begin{table}[t]
\footnotesize
\caption{\textbf{Comparison of emotion-based datasets of faces in challenging conditions.}}\label{t:datasets}\label{table:headings}
\begin{center}
\setlength{\tabcolsep}{4pt}
\begin{tabular}{lrlll}
\hline\noalign{\smallskip}
Dataset & Size\;\;\;\;\;\;\;\;\; & Labelling Technique & Expressions & Labels \\
\noalign{\smallskip}
\hline
\noalign{\smallskip}
FER                & 35,887 Faces     &    Internet search  & Mixed       & 6+1 emotions \\
AFEW 5.0           & 1,426 Clips      &    Subtitles               & Acted       & 6+1 emotions \\
SFEW 2.0         & 1,635 Faces     &    Subtitles               & Acted       & 6+1 emotions \\
AM-FED             & 168,359 Faces   &    Human experts           & Spontaneous & FACS \\
\dataset (ours)    & 192,030 Faces   &   Metadata extraction     & Spontaneous & Event Outcome \\
\hline
\end{tabular}
\end{center}
\vspace{-2em}
\end{table}

Facial expressions are a non-verbal mode of communication complementary to speech and gestures~\cite{ekman1969repertoire,attardo2003multimodal}. They can  be produced unintentionally~\cite{ekman1969nonverbal}, revealing hidden states of the actor in pain or deception detection~\cite{besel2010individual}. Facial expressions are commercially valuable, attracting increasing investment from advertising agencies that seek to understand and manipulate the consumer response to a product~\cite{el2012affect} and corresponding regulatory attention~\cite{CapuanoWAWYA2015}.

Face-related tasks such as face detection, verification and recognition have long been researched in computer vision with the creation of several labelled datasets: FDDB~\cite{jain2010fddb}, AFW~\cite{zhu2012face} and AFLW~\cite{koestinger11annotated} for face detection; and LFW~\cite{huang2007labeled} and VGG-Face~\cite{Parkhi15} for face recognition and verification. Face detectors and identity recognizers can now rival the performance of humans~\cite{schroff2015facenet}. Facial expression recognition has also received significant attention in computer vision, but it presents a number of additional subtleties and difficulties which are not found in face detection or recognition. The main challenge is the consistent labelling of facial expressions which is difficult due to the subjective nature of the task. A number of coding systems have been developed in an attempt to label facial expressions objectively, usually at the level of atomic facial movements, but even human experts are not infallible in generating such annotations. Furthermore, getting these experts to annotate a dataset is expensive and difficult to scale~\cite{mcduff2013affectiva}. Another issue is the ``authenticity'' of facial expressions, arising from the fact that several datasets are acted~\cite{sebe2007authentic}, either specifically for data collection~\cite{lyons1998coding}~\cite{lucey2010extended}~\cite{gross2010multi} or indirectly as data is extracted from movies~\cite{dhall15video}. Our \dataset dataset sidesteps these problems by recording spontaneous reactions to objectively-occurring events in videos.

Examples of datasets which contain challenging variations in pose, lighting conditions and subjects are given in Table~\ref{t:datasets}. Of these, two in particular have received significant research interest as popular benchmarks for facial expression recognition. The \emph{Static Facial Expression in the Wild 2.0} (SFEW-2.0) data~\cite{dhall11static} (used in the ~\emph{EmotiW} challenges~\cite{dhall15video}) consists of images from movies which collectively contain 1,635 faces labelled with seven emotions (this dataset was constructed by selectively extracting individual frames from AFEW-5.0~\cite{dhall2012collecting}). The \emph{Facial Expression Recognition 2013} (FER-2013) dataset~\cite{goodfellow15challenges}, which formed the basis of a large Kaggle competition, contains 35k images labelled with the same seven emotions. These datasets were used to develop several state-of-the-art emotion recognition systems. Among the top-performing ones, the authors of~\cite{yu2015image} and~\cite{kim2016hierarchical} propose ensembles of deep network trained on the FER and SFEW-2.0 data. There are also several commercial implementations of expression recognition, such as CMU's IntraFace~\cite{de2015intraface} and the Affectiva face software.

% ---------------------------------------------------------
\section{\dataset: expressions in context}\label{s:data}
% ---------------------------------------------------------

In this section we describe the \dataset dataset (Fig.~\ref{f:splash}) and how it was collected.

\paragraph{Data source.} The ``Deal or No Deal'' TV game show\footnote{Outside of computer vision, the interesting decision making dynamics of contestants in a high-stakes environment during the ``Deal or No Deal'' game show have attracted research by economists \cite{post2008deal}.} was selected as the basis for our data for a number of reasons. First, it contains a very significant amount of data. The show has been running nearly daily in the UK for the past eleven years, totalling 2,929 episodes. Each episode focuses on a different player and lasts for about forty minutes. Furthermore, the same or very similar shows are or were aired in dozens of other countries. Second, the game is based on simple rules and a sequence of discrete events that are in most cases easily identifiable as positive or negative for the player, and hence can be expected to induce a corresponding emotion and facial expression. Furthermore, these events are easily detectable by parsing textual overlays in the show or other simple patterns. Thirdly, since there is a single player, it is easy to identify the person that is directly affected by the events in the video and the camera tends to focus on his/her face.

An example of the in-game footage and data extraction pipeline is shown in Fig.~\ref{f:splash}. The rules of the game are easily explained. There are $n=22$ possible cash prizes $\mathcal{X}_0 = \{p_1,p_2,\dots,p_n\}$ where prizes $p_1 < p_2 < \dots < p_n$ range from 1p up to \pounds 250,000. Initially the player is assigned a prize $x_0 \in \mathcal{X}_0$ but does not know its value. Then, at each round of the game the player can randomly extract (realised as opening a box, see Fig.~\ref{f:splash} top-left) one of the prizes $x_t \not = x_0$ from $\mathcal{X}_t$ and reveal it, resulting in a smaller set $\mathcal{X}_t = \mathcal{X}_{t-1} - \{x_t\}$ of possible prizes. Through this process of elimination the player obtains information about his/her prize $x_0$. Occasionally the player is offered the opportunity to leave the game with a prize $p_d$ (``deal'') determined by the game's host or to continue playing (``no deal'') and eventually leave with $x_0$.

The expected value $E_t$ of the win $x_0$ at time $t$ is $E_t = \operatorname{mean} \mathcal{X}_t$. When a prize $x_{t}$ is removed from $\mathcal{X}_{t-1}$, the player perceives this as a ``good'' event if $E_{t} > E_{t-1}$, which requires $x_{t} < E_{t-1}$, and a ``bad'' event otherwise.  In practice we conservatively require $E_t > E_{t-1} + \Delta$ for a good event, where $\Delta = \pounds 750$. Interestingly, the game is continued even after the player has taken a ``deal''; in this case the roles of ``good'' and ``bad'' events are reversed as the player hopes that the accepted deal $p_d$ is higher than the prize $x_0$ he/she gave up.

\paragraph{Dataset content.} The data in \dataset is defined as follows. Faces are detected right after a new prize $x_t$ is revealed for about seven seconds. These faces are collected in a ``face track'' $\bff_t$. Furthermore, the face track is assigned the binary label:
\[
 y_t = 
 d_t
 \times
 \begin{cases}
 +1, & x_{t} + \Delta < E_{t-1}, \\
 -1, & x_{t} + \Delta \geq E_{t-1},
 \end{cases} 
\]
where $d_t$ is $+1$ if the deal was \emph{not} taken so far, and $-1$ otherwise. Note that there are several levels of indirection between $y_t$ and a particular expression being shown in $\bff_t$. For example, a player may not perceive a good or bad event according to this simple model, or could be responding to a stroke of bad luck with an ironic smile. The labels  $y_t$ themselves, however, are completely objective.

Data is extracted from 102 episodes of the show, resulting in 192,030 frames distributed over 2,118 labelled face tracks. Shows are divided into training, validation and test sets, which also means that mostly different identities are contained in the different subsets. 

\paragraph{Data extraction.} One advantage of studying facial expressions from contextual events is that these are often easy to detect automatically. In our case, we take advantage of two facts. First, when  a prize is removed from the set $\mathcal{X}_t$, this is shown in the game as a box being opened (Fig.~\ref{f:splash} top-left). This scene, which occurs systematically, is easy to detect and is used to mark the start of an event. Next, the camera moves onto the contestant (Fig.~\ref{f:splash} top-middle) to capture his/her reaction. Faces are extracted from the seven seconds that immediately follow the event using the face detector of~\cite{dlib09} and are stored as part of the face track $\bff = (f_1,f_2,\dots,f_T)$. Occasionally the camera may capture the reaction of a member of the public; while it would be easy to distinguish different identities (e.g.\ by using the VGG-Faces model of Sect.~\ref{s:networks}), we prefer not to as the public is sympathetic with the contestant and tends to react in a similar manner, improving the diversity of the collected data. Finally, the value of the prize $x_t$ being removed can be extracted either from the opened box using a text spotting system or, more easily, by looking at which overlay is removed (Fig.~\ref{f:splash} top-right). After automatic extraction, the data was fully checked manually for errors to ensure its quality.

% ---------------------------------------------------------
\section{Benchmark data and human baselines}\label{s:human}
% ---------------------------------------------------------

As \dataset defines a new task in facial expression interpretation, in this section we establish a human baseline as a point of comparison with computer vision algorithm performance. In order to compare \dataset to existing facial expression recognition problems we establish similar baselines for two standard expression recognition datasets, FER and SFEW 2.0, introduced below.

\paragraph{Benchmark datasets: FER and SFEW 2.0.} The FER-2013 data~\cite{goodfellow15challenges} contains $48\times 48$ pixel images obtained by querying Google image search for 184 emotion-related keywords. The dataset contains 35,887 images divided into  4,953 ``anger'', 547 ``disgust'', 5,121 ``fear'', 8,989 ``happiness'', 6,077 ``sadness'', 4,002 ``surprise'' and 6,198 ``neutral'' further split into training (28,709), public test (3,589) and private test (3,589) sets. Goodfellow~\etal~\cite{goodfellow15challenges} note that this data is likely to contain label errors. However, their own human study obtained an average prediction accuracy of $65\pm 5\%$, which is comparable to the $68 \pm 5\%$ performance obtained by expert annotators on a smaller but manually-curated subset of 1,500 acted images.

The SFEW-2.0 data~\cite{dhall11static} contains selected frames from different videos of the \emph{Acted Facial Expressions in the Wild} (AFEW) dataset~\cite{dhall11acted} assigned to either: 225 ``angry'', 75 ``disgust'', 124 ``fear'', 256 ``happy'', 228 ``neutral'', 234 ``sad''  and 150 ``surprise''. The training, validation and test splits are provided as part of the EmotiW challenge~\cite{dhall15video} and are adopted here. The AFEW data was collected by searching movie close captions for emotion-related keywords and then manually curating the results, generating a smaller number of labelled instances than FER.

\paragraph{Human baselines.} For each dataset we consider a pool of annotators, most of which are not computer vision experts, and ask them to predict the label associated with each face. In order to motivate annotators to be as accurate as possible, we pose the annotation process as a challenge. The goal is to guess the ground-truth label of an image and a score displaying the annotators' prediction accuracy is constantly updated. Ultimately, annotators performances are entered in a leaderboard. We found that this simple idea significantly improved the annotators' performance. 

The dataset instances selected for the annotation tasks were constructed as follows. From FER, a random sample of 500 faces was extracted from the Public Test set. From SFEW 2.0, the full Validation set (383 samples) was used (faces were extracted from each image as described in section~\ref{s:networks}). From \dataset, a random sample of 250 face tracks was extracted from the validation set, each of which was transformed into an animated GIF to allow annotators to see the face motion. Performance on each dataset was evaluated by partitioning into five folds, each of which was annotated by a separate pool. Every face instance across the three datasets received at least four annotations.

 On FER, our annotators achieved lower performance than results previously reported in~\cite{goodfellow15challenges} (58.2\% overall accuracy vs 65\%). However, we also noted a significant variance between annotators ($\pm 8.0\%$), which means that at least some of them were able to match or exceed the $65\%$ mark. The unevenness of the annotators shows how difficult or ambiguous this task can be even for motivated humans. The annotators found SFEW-2.0 a more challenging task, obtaining an average accuracy of $53.0\pm 9.4\%$ overall.  One possible reason for this difference is the manner in which the datasets were constructed. FER faces were retrieved using Internet search queries which likely returned fairly representative examples of each expression; in contrast SFEW images were extracted from movies. On \dataset, the average annotator accuracy was $62.0\pm8.1\%$. Since the classification task was binary, to facilitate a comparison with algorithmic approaches, the ROC-AUC was also computed for each annotator, resulting in an annotator average of $71.0\pm5\%$. The relatively low scores of humans on each dataset illustrate the particularly challenging nature of the task. This difficulty is underlined by the low levels of inter-annotator agreement (measured using \textit{Fleiss' kappa}) on the three datasets of 0.574, 0.424 and 0.491 respectively.   

% ---------------------------------------------------------
\section{Expression recognition networks}\label{s:networks}
% ---------------------------------------------------------

In this section we develop state-of-the-art models for facial expression recognition in the two popular emotion recognition benchmarks of Sect.~\ref{s:human}, namely FER and SFEW 2.0. Deep networks are currently the state-of-the-art models for emotion recognition, topping two of the last three editions of the \textit{Emotion recognition in the Wild} (EmotiW) contest~\cite{levi15emotion}. While the standard approach is to learn large ensembles of deep networks~\cite{kim2016hierarchical,yu2015image}, here we show that a single network can in fact be competitive or better than such ensembles if trained effectively. In order to do so we expand the available training data by pre-training models on other face recognition tasks, and in particular face identity verification, using the recent VGG-Faces dataset~\cite{parkhi15deep}.

\paragraph{Architectures and training.} We base our models on four standard CNN architectures: AlexNet~\cite{krizhevsky2012imagenet}, VGG-M~\cite{chatfield14return}, VGG-VD-16~\cite{simonyan15very} and ResNet-50~\cite{he2016deep}. AlexNet is used as a reference baseline and is pre-trained on the ImageNet ILSVRC data~\cite{russakovsky14imagenet}. VGG-VD-16 is pre-trained on a recent dataset for face verification called VGG-Faces~\cite{parkhi15deep}. This model achieves near state-of-the-art verification performance on the LFW~\cite{huang2007labeled} benchmark; however, it is also extremely expensive. Thus, we train also a smaller network, based on the VGG-M configuration. All models are trained with batch normalization~\cite{ioffe15batch} and are implemented in the MatConvNet framework \cite{vedaldi2015matconvnet}.

Statistics such as image resolution and the usage of colour in the target datasets, and FER in particular, differ substantially from LFW and VGG-Faces. Nevertheless, we found that simply rescaling the smaller FER images to the higher VGG-Faces resolution together with duplicating the grayscale intensities for the three colour channels produced excellent results. We also experimented with the other approach of pretraining by reducing the resolution and removing colour information from VGG-Faces; while this resulted in very competitive and more efficient networks, the full resolution models were still a little more accurate and are used in the rest of the work.

After pre-training, each model is trained on the FER or SFEW 2.0 training set with a fine tuning ratio of 0.1. This is obtained by retaining all but the last layer, performing $N$-way classification, where $N$ is the number of possible facial expression classes.

\begin{figure}
\newcommand{\putim}[2]{%
\begin{overpic}[width=0.333\linewidth]{#1}
\put (0,88 ) {{\colorbox{white}{\tiny #2}}}
\end{overpic}}
\footnotesize
\setlength{\tabcolsep}{2pt}
\begin{floatrow}
\capbtabbox{%
\caption{Accuracy on FER-2013 of different CNN models and training strategies.}\label{t:fer}%
}{%
\begin{tabular}{llllll}
\hline\noalign{\smallskip}
Model & Pretraining & Test (Public) & Test (Private) \\
\noalign{\smallskip}
\hline
\noalign{\smallskip}
AlexNet           & ImageNet &    62.44\%             &    63.28\%             \\
VGG-M             & ImageNet &    66.04\%             &    67.57\%             \\
Resnet-50         & ImageNet &    67.79\%             &    69.02\%             \\
VGG-VD-16         & ImageNet &    66.92\%             &    70.38\%             \\

\hline
\noalign{\smallskip}
AlexNet           & VGGFaces &    70.47\%            &    71.44\%              \\
VGG-M             & VGGFaces &    71.08\%            &    72.08\%              \\
Resnet-50         & VGGFaces &    69.23\%            &     70.33\%              \\   
VGG-VD-16         & VGGFaces &   \textbf{72.05}\%    & \textbf{72.89}\%        \\  

\hline
\hline
$\text{HDC}\star$~\cite{kim2016hierarchical}     &   -       &       -         &     70.58\%   \\
$\text{HDC}\dagger\dagger$~\cite{kim2016hierarchical}                &   -       &       -         &     72.72\%    \\
\hline
\end{tabular}%
}

\capbtabbox{%
\caption{Accuracy on SFEW-2.0 of different CNN models and training strategies}\label{t:sfew}%
}{%
\label{table:fer}
\begin{tabular}{llllll}
\hline\noalign{\smallskip}
Model & Pretraining & Val & Test \\
\noalign{\smallskip}
\hline
\noalign{\smallskip}
AlexNet           & VGGFaces &    37.67\%            &    -             \\
VGG-M             & VGGFaces &    42.90\%            &    -             \\
Resnet-50         & VGGFaces &    47.48\%            &    -             \\
VGG-VD-16         & VGGFaces &    43.58\%            &    -             \\
\hline
\noalign{\smallskip}
AlexNet           & VGGFaces+FER &    38.07\%            &    50.81\%            \\
VGG-M             & VGGFaces+FER &    47.02\%            &    53.49\%            \\
Resnet-50         & VGGFaces+FER &    50.91\%            &    45.97\%            \\
VGG-VD-16         & VGGFaces+FER &\bf 54.82\%            &\underline{59.41}\%    \\
\hline
\hline
$CMU\star$~\cite{yu2015image}             &   FER combined &       52.29\%    &     58.06\%          \\
$\text{HDC}\star$~\cite{kim2016hierarchical} &   FER + TFD    &       52.50\%     &     57.3\%           \\
\hline
$CMU\dagger\dagger$~\cite{yu2015image}               &   FER combined &       55.96\%    &     61.29\%          \\
$\text{HDC}\dagger\dagger$~\cite{kim2016hierarchical}            &   FER + TFD    &       52.80\%     &     \textbf{61.6\%}  \\
\hline
\end{tabular}
}

\end{floatrow}
\end{figure}

\begin{figure}
\newcommand{\putim}[2]{%
\begin{minipage}[t]{0.139\linewidth}%
\centering\includegraphics[width=\linewidth,height=\linewidth]{#1}\\
\footnotesize #2%
\end{minipage}}
\begin{center}
\putim{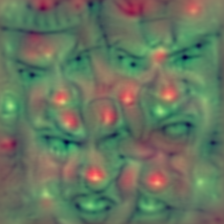}{Anger}\,%
\putim{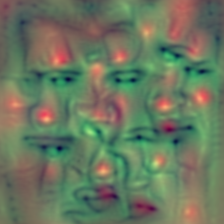}{Disgust}\,%
\putim{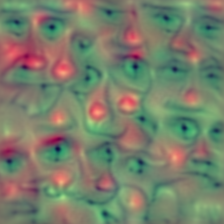}{Fear}\,%
\putim{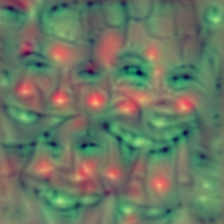}{Happiness}\,%
\putim{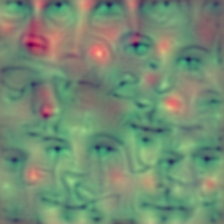}{Neutral}\,%
\putim{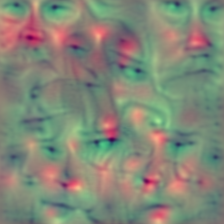}{Sadness}\,%
\putim{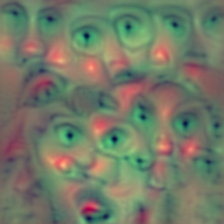}{Surprise}
\end{center}
\vspace{-2em}
\caption{Visualizations of the FER emotions for the VGG-VD-16 model.}\label{f:vis}
\end{figure}

\paragraph{Results.} Table~\ref{t:fer} compares the different architecture and the state-of-the-art on FER. When reporting ensemble models, $\star$ denotes the best single CNN and $\dagger\dagger$ denotes the ensemble. The best previous results on FER is 72.72\% accuracy, obtained using the hierarchical committee of deep CNNs described in \cite{kim2016hierarchical}, combining more than 36 different models. By comparison, VGG-VD-16 pre-trained on VGG-Faces achieves a slightly superior performance at 72.89\%. VGG-M achieves nearly the same performance ($-0.8\%$) at a substantially reduced computational cost. We also note the importance of choosing a face-related pre-training set, as pre-training in ImageNet loses 3-4\% of performance.

Table~\ref{t:sfew} reports the results on the SFEW-2.0 dataset instead. Since the dataset itself consists of labelled scene images, we use the faces extracted by the accurate face detection pipeline described in \cite{yu2015image} which applies an ensemble of face detectors \cite{zhang2014improving,chen2014joint, zhu2012face}. As SFEW is much smaller than FER, pre-training is in this case much more important.  The best result achieved by any of the four models pre-trained with ImageNet only was $31.19\%$. Pre-training on VGG-Faces produced substantially better results (+10\%), and pre-training on VGG- Faces and FER-Train produced better still (+18\%). The best single model, VGG-VD-16, achieves better performance than existing single and ensemble networks (+2.5\%) on the validation set, and better performance than all but the ensembles of~\cite{yu2015image,kim2016hierarchical} on the test set (-2\%).

\paragraph{Visualizations.} While CNNs perform well, it is often difficult to understand what they are learning given their black-box nature. Here we use the technique of~\cite{mahendran16visualizing} to visualize the the best FER/SFEW model. This technique seeks to find an image $I$ which, under certain regularity assumptions, maximizes the CNN confidence $\Phi_c(I)$ that $I$ represents emotion $c$. Results are reported in Fig~\ref{f:vis} for the VGG-VD-16 model trained on the FER dataset. Notably, the reconstructed pictures are mosaics of parts representative of the corresponding emotions.

% ---------------------------------------------------------
\section{Relating facial expressions to events in videos}\label{s:context}
% ---------------------------------------------------------

In this section we focus on the main question of the paper i.e.\  whether facial expressions can be used to extract information about events in videos.

\paragraph{Baselines: individual frame prediction and simple voting.} As baseline, a state-of-the-art emotion recognition CNN $\Phi$ is applied to each frame in the face track. The $T$ faces in a face track $\bff=(f_1,\dots,f_T)$ are individually classified by $\Phi(f_t)$ and results are pooled to predict whether the event is positive $y = +1$ or negative $y = -1$. Positive emotions (happiness) vote for the first case, negative emotions (sadness, fear, anger, disgust) for the second and neutral/surprise emotions are ignored. The label with the largest number of votes in the track wins.

\paragraph{Pooling architectures.} There are two significant shortcomings in the baseline. First, it assumes a particular map between emotions in existing datasets and positive and negative events in \dataset. Second, it integrates information across frames using an ad-hoc voting procedure which may be suboptimal. In order to address these shortcomings we learn on \dataset a new model that explicitly pools information across frames in a track. A pre-trained network $\Phi = \Phi_1 \circ \Phi_2$ is split in two parts. Then, the first part is run independently on each frame, the results are pooled by either average or max pooling across time and the result is fed to $\Phi_2$ for binary classification: $\Phi(\bff) = \Phi_2 \circ \operatorname{pool}(\Phi_1(f_1),\dots,\Phi_1(f_T))$. The resulting architecture is fine-tuned on the \dataset training set.

In practice, we found that the best results were obtained by using the emotion recognition networks such as VGG-VD-16 trained on the FER data (Sect.~\ref{s:networks}). All layers up to fc7, producing 4,096 dimensional feature vectors, are retained in $\Phi_1$. The best pooling function was found to be averaging followed by $L^1$ normalization of the 4,096 dimensional features. The last layer $\Phi_8$ is fully connected (in practice, this layer is a linear predictor). CNNs are trained using hinge loss, which generally performs better than softmax for binary classification.

\begin{table}[t]
\footnotesize
\setlength{\tabcolsep}{3pt}
\begin{floatrow}
\capbtabbox{%
\caption{ROC-AUC on \dataset}\label{t:results}%
}{%
\label{table:fer}
\begin{tabular}{llllll}
\hline\noalign{\smallskip}
Model & Pre-training & Method & Val. & Test\\
\noalign{\smallskip}
\hline
\noalign{\smallskip}
 VGG-M         &  VGGFace+FER &  voting      & 0.656          &  0.592             \\
 VGG-VD      &  VGGFace+FER &  voting      & 0.653          &  0.618             \\
\hline
\noalign{\smallskip}
 VGG-M         &  VGGFace & pooling arch.    & 0.764         &   0.691             \\
 VGG-VD     &  VGGFace & pooling arch.   & 0.726         &   0.671              \\

 \hline
\noalign{\smallskip}
 VGG-M         &  VGGFace+FER & pooling arch.    & \textbf{0.794} &  \textbf{0.722}        \\
 VGG-VD     &  VGGFace+FER & pooling arch.     & 0.741          &     0.675              \\
\hline
\end{tabular}
}
\ffigbox{%
\caption{FER expressions from \dataset.}\label{f:learning-emotions}}{%
\includegraphics[width=0.35\textwidth]{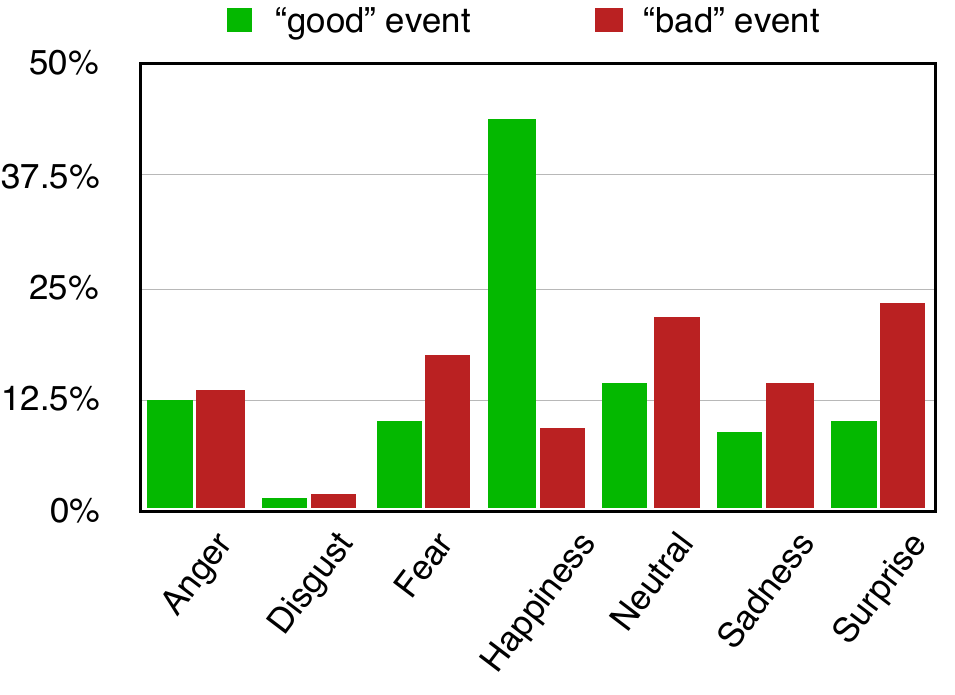}\vspace{-1.5em}%
}
\end{floatrow}
\end{table}

\begin{table}[t]
\footnotesize
\setlength{\tabcolsep}{5pt}
\ttabbox[\linewidth]{%
\caption{Comparison of human vs machine performance across benchmarks}\label{t:humanVmachine}%
}{%
\label{t:vshumans}
\begin{tabular}{llllll}
\hline\noalign{\smallskip}
Dataset & Metric & Human & Human Committee & Machine\\
\noalign{\smallskip}
\hline
\noalign{\smallskip}
 FER (public test)      &   Accuracy      & 0.57     & 0.66   &   0.72                        \\
 SFEW 2.0 (val)         &   Accuracy      & 0.53     & 0.63   &   0.56 \cite{yu2015image}     \\
 \dataset (val)         &   ROC-AUC       & 0.71     & 0.78 &   0.79                       \\
\hline

\end{tabular}
}
\end{table}

\paragraph{Results.} Table~\ref{t:results} reports the performance of different model variants on \dataset. Similarly to Table~\ref{t:sfew}, pre-training on VGG-Face+FER is preferable than pre-training on VGG-Face only. This is required for the voting classifier, but beneficial also when fine-tuning a pre-trained pooling architecture, which handily outperforms voting. VGG-M is in this case better than VGG-VD ($+5.3\%$), probably due to the fact that VGG-VD is overfitted to the pre-training data. Finally, temporal average pooling is always better than max pooling.

\paragraph{Learning nameable facial expressions from events in videos.} So far, we have shown that it is possible to predict events in videos by looking at facial expressions. Here we consider the other direction and ask whether nameable facial expressions can be learned by looking at people in TV programs reacting to events. To answer this question we applied the VGG-M pooling architecture to the FER images after pre-trained it on VGG-Faces (a verification task) and fine-tuning it on \dataset. In this manner, this CNN is never trained with manually-labelled emotions. Fig.~\ref{f:learning-emotions} shows the distribution of FER nameable expressions for faces associated to ``good'' and ``bad'' \dataset events by this model. There is a marked difference in the resulting distributions, with a significant peak for \textit{happiness} for predicted ``good'' events and \textit{surprise} and negative emotions for ``bad'' ones. This suggests that it is indeed possible to learn nameable expressions from their weak association to events in video without explicit and dedicated supervision as commonly done.

\paragraph{Comparison with human baselines.} Table~\ref{t:vshumans} compares the performance of humans and of the best models on the three datasets FER, SFEW 2.0, and \dataset. Remarkably, in all cases networks outperform individual humans by a substantial margin (\eg +15\% on FER and +8\% on \dataset). While this result is perhaps surprising, we believe the reason is that, in such ambiguous tasks, machines learn to respond as humans would on \emph{average} whereas the performance of \emph{individual} annotators, as reflected in Table~\ref{t:vshumans}, can be low due to poor inter-annotator agreement. To verify this hypothesis, we combined multiple human annotators in a committee and found that this gap either closes or disappears. In particular, on \dataset the performance of the committee is just a hair's breadth lower than that of the machine (78\% vs 79\%).

% ---------------------------------------------------------
\section{Summary}\label{s:summary}
% ---------------------------------------------------------

In this paper we have investigated the problem of relating facial expressions with objectively-measurable events that affect humans in videos. We have shown that gameshows are a particularly useful data source for this type of analysis due to their simple structure, easily detectable events and emotional impact on the participants and have constructed a corresponding dataset \dataset.

In order to analyze emotions in \dataset, we have trained state-of-the-art neural networks for facial expression recognition in existing datasets showing that, if pre-trained on face verification, single models are competitive or better than the multi-network committees commonly used in the literature. Then, we have shown that such networks can successfully understand the relationship between certain events in TV programs and facial expressions better than individual human annotators, and as well as a committee of several human annotators. We have also shown that networks trained to predict such events from facial expressions correlate very well to nameable expressions in standard datasets.
% ---------------------------------------------------------
\subsubsection*{Acknowledgements}\label{s:acknowledgements}
% ---------------------------------------------------------

The authors gratefully acknowledge the support of the ESPRC EP/L015897/1 (AIMS CDT) and the ERC 677195-IDIU. We also wish to thank Zhiding Yu for kindly sharing his preprocessed SFEW dataset.

% ---------------------------------------------------------
\bibliography{egbib}

\begin{thebibliography}{39}
\providecommand{\natexlab}[1]{#1}
\providecommand{\url}[1]{\texttt{#1}}
\expandafter\ifx\csname urlstyle\endcsname\relax
  \providecommand{\doi}[1]{doi: #1}\else
  \providecommand{\doi}{doi: \begingroup \urlstyle{rm}\Url}\fi

\bibitem[Attardo et~al.(2003)Attardo, Eisterhold, Hay, and
  Poggi]{attardo2003multimodal}
Salvatore Attardo, Jodi Eisterhold, Jennifer Hay, and Isabella Poggi.
\newblock Multimodal markers of irony and sarcasm.
\newblock \emph{Humor}, 16\penalty0 (2):\penalty0 243--260, 2003.

\bibitem[Besel and Yuille(2010)]{besel2010individual}
Lana~DS Besel and John~C Yuille.
\newblock Individual differences in empathy: The role of facial expression
  recognition.
\newblock \emph{Personality and Individual Differences}, 49\penalty0
  (2):\penalty0 107--112, 2010.

\bibitem[Chatfield et~al.(2014)Chatfield, Simonyan, Vedaldi, and
  Zisserman]{chatfield14return}
K.~Chatfield, K.~Simonyan, A.~Vedaldi, and A.~Zisserman.
\newblock Return of the devil in the details: Delving deep into convolutional
  nets.
\newblock 2014.

\bibitem[Chen et~al.(2014)Chen, Ren, Wei, Cao, and Sun]{chen2014joint}
Dong Chen, Shaoqing Ren, Yichen Wei, Xudong Cao, and Jian Sun.
\newblock Joint cascade face detection and alignment.
\newblock In \emph{European Conference on Computer Vision}, pages 109--122.
  Springer, 2014.

\bibitem[de~la Torre et~al.(2015)de~la Torre, Chu, Xiong, Vicente, Ding, and
  Cohn]{de2015intraface}
Fernando de~la Torre, Wen-Sheng Chu, Xuehan Xiong, Francisco Vicente, Xiaoyu
  Ding, and Jeffrey Cohn.
\newblock Intraface.
\newblock In \emph{Automatic Face and Gesture Recognition (FG), 2015 11th IEEE
  International Conference and Workshops on}, volume~1, pages 1--8. IEEE, 2015.

\bibitem[Dhall et~al.(2011{\natexlab{a}})Dhall, Goecke, Lucey, and
  Gedeon]{dhall11acted}
A.~Dhall, R.~Goecke, S.~Lucey, and T.~Gedeon.
\newblock {Acted Facial Expressions in the Wild Database}.
\newblock Technical report, Australian National University, 2011{\natexlab{a}}.

\bibitem[Dhall et~al.(2011{\natexlab{b}})Dhall, Goecke, Lucey, and
  Gedeon]{dhall11static}
Abhinav Dhall, Roland Goecke, Simon Lucey, and Tom Gedeon.
\newblock {Static facial expression analysis in tough conditions: Data,
  evaluation protocol and benchmark}.
\newblock In \emph{Proc. {ICCV} Workshop}, 2011{\natexlab{b}}.

\bibitem[Dhall et~al.(2015)Dhall, Ramana~Murthy, Goecke, Joshi, and
  Gedeon]{dhall15video}
Abhinav Dhall, O.V. Ramana~Murthy, Roland Goecke, Jyoti Joshi, and Tom Gedeon.
\newblock Video and image based emotion recognition challenges in the wild:
  Emotiw 2015.
\newblock In \emph{Proc. {ACM} Int. Conf. on Multimodal Interaction}, 2015.

\bibitem[Dhall et~al.(2012)]{dhall2012collecting}
Abhinav Dhall et~al.
\newblock Collecting large, richly annotated facial-expression databases from
  movies.
\newblock 2012.

\bibitem[Ekman and Friesen(1969{\natexlab{a}})]{ekman1969nonverbal}
Paul Ekman and Wallace~V Friesen.
\newblock Nonverbal leakage and clues to deception.
\newblock \emph{Psychiatry}, 32\penalty0 (1):\penalty0 88--106,
  1969{\natexlab{a}}.

\bibitem[Ekman and Friesen(1969{\natexlab{b}})]{ekman1969repertoire}
Paul Ekman and Wallace~V Friesen.
\newblock The repertoire of nonverbal behavior: Categories, origins, usage, and
  coding.
\newblock \emph{Semiotica}, 1\penalty0 (1):\penalty0 49--98,
  1969{\natexlab{b}}.

\bibitem[El~Kaliouby et~al.(2012)El~Kaliouby, Dreisch, England, and
  Kodra]{el2012affect}
Rana El~Kaliouby, Andrew~Edwin Dreisch, Avril England, and Evan Kodra.
\newblock Affect based concept testing, December~27 2012.
\newblock US Patent App. 13/728,303.

\bibitem[Goodfellow et~al.(2015)Goodfellow, Erhan, Carrier, Courville, Mirza,
  Hamner, Cukierski, Tang, Thaler, Lee, Zhou, Ramaiah, Feng, Li, Wang,
  Athanasakis, Shawe-Taylor, Milakov, Park, Ionescu, Popescu, Grozea, Bergstra,
  Xie, Romaszko, Xu, Chuang, and Bengio]{goodfellow15challenges}
Ian~J. Goodfellow, Dumitru Erhan, Pierre~Luc Carrier, Aaron Courville, Mehdi
  Mirza, Ben Hamner, Will Cukierski, Yichuan Tang, David Thaler, Dong-Hyun Lee,
  Yingbo Zhou, Chetan Ramaiah, Fangxiang Feng, Ruifan Li, Xiaojie Wang,
  Dimitris Athanasakis, John Shawe-Taylor, Maxim Milakov, John Park, Radu
  Ionescu, Marius Popescu, Cristian Grozea, James Bergstra, Jingjing Xie,
  Lukasz Romaszko, Bing Xu, Zhang Chuang, and Yoshua Bengio.
\newblock Challenges in representation learning: A report on three machine
  learning contests.
\newblock \emph{Neural Networks}, 64:\penalty0 59 -- 63, 2015.

\bibitem[Gross et~al.(2010)Gross, Matthews, Cohn, Kanade, and
  Baker]{gross2010multi}
Ralph Gross, Iain Matthews, Jeffrey Cohn, Takeo Kanade, and Simon Baker.
\newblock Multi-pie.
\newblock \emph{Image and Vision Computing}, 28\penalty0 (5):\penalty0
  807--813, 2010.

\bibitem[He et~al.(2016)He, Zhang, Ren, and Sun]{he2016deep}
Kaiming He, Xiangyu Zhang, Shaoqing Ren, and Jian Sun.
\newblock Deep residual learning for image recognition.
\newblock In \emph{Computer Vision and Pattern Recognition (CVPR), 2016 IEEE
  Conference on}, 2016.

\bibitem[Huang et~al.(2007)Huang, Ramesh, Berg, and
  Learned-Miller]{huang2007labeled}
Gary~B Huang, Manu Ramesh, Tamara Berg, and Erik Learned-Miller.
\newblock Labeled faces in the wild: A database for studying face recognition
  in unconstrained environments.
\newblock Technical report, Technical Report 07-49, University of
  Massachusetts, Amherst, 2007.

\bibitem[Ioffe and Szegedy(2015)]{ioffe15batch}
S.~Ioffe and C.~Szegedy.
\newblock Batch normalization: Accelerating deep network training by reducing
  internal covariate shift.
\newblock \emph{CoRR}, 2015.

\bibitem[Jain and Learned-Miller(2010)]{jain2010fddb}
Vidit Jain and Erik~G Learned-Miller.
\newblock Fddb: A benchmark for face detection in unconstrained settings.
\newblock \emph{UMass Amherst Technical Report}, 2010.

\bibitem[Kim et~al.(2016)Kim, Roh, Dong, and Lee]{kim2016hierarchical}
Bo-Kyeong Kim, Jihyeon Roh, Suh-Yeon Dong, and Soo-Young Lee.
\newblock Hierarchical committee of deep convolutional neural networks for
  robust facial expression recognition.
\newblock \emph{Journal on Multimodal User Interfaces}, pages 1--17, 2016.

\bibitem[King(2009)]{dlib09}
Davis~E. King.
\newblock Dlib-ml: A machine learning toolkit.
\newblock \emph{Journal of Machine Learning Research}, 10:\penalty0 1755--1758,
  2009.

\bibitem[Koestinger et~al.(2011)Koestinger, Wohlhart, Roth, and
  Bischof]{koestinger11annotated}
Martin Koestinger, Paul Wohlhart, Peter~M. Roth, and Horst Bischof.
\newblock Annotated facial landmarks in the wild: A large-scale, real-world
  database for facial landmark localization.
\newblock In \emph{{ICCV} Workshop on Benchmarking Facial Image Analysis
  Technologies}, 2011.

\bibitem[Krizhevsky et~al.(2012)Krizhevsky, Sutskever, and
  Hinton]{krizhevsky2012imagenet}
Alex Krizhevsky, Ilya Sutskever, and Geoffrey~E Hinton.
\newblock Imagenet classification with deep convolutional neural networks.
\newblock In \emph{Advances in neural information processing systems}, pages
  1097--1105, 2012.

\bibitem[Levi and Hassner(2015)]{levi15emotion}
Gil Levi and Tal Hassner.
\newblock Emotion recognition in the wild via convolutional neural networks and
  mapped binary patterns.
\newblock In \emph{Proc. {ACM} Int. Conf. on Multimodal InteractionP}, 2015.

\bibitem[Lucey et~al.(2010)Lucey, Cohn, Kanade, Saragih, Ambadar, and
  Matthews]{lucey2010extended}
Patrick Lucey, Jeffrey~F Cohn, Takeo Kanade, Jason Saragih, Zara Ambadar, and
  Iain Matthews.
\newblock The extended cohn-kanade dataset (ck+): A complete dataset for action
  unit and emotion-specified expression.
\newblock In \emph{Computer Vision and Pattern Recognition Workshops (CVPRW),
  2010 IEEE Computer Society Conference on}, pages 94--101. IEEE, 2010.

\bibitem[Lyons et~al.(1998)Lyons, Akamatsu, Kamachi, and
  Gyoba]{lyons1998coding}
Michael Lyons, Shota Akamatsu, Miyuki Kamachi, and Jiro Gyoba.
\newblock Coding facial expressions with gabor wavelets.
\newblock In \emph{Automatic Face and Gesture Recognition, 1998. Proceedings.
  Third IEEE International Conference on}, pages 200--205. IEEE, 1998.

\bibitem[Mahendran and Vedaldi(2016)]{mahendran16visualizing}
Aravindh Mahendran and Andrea Vedaldi.
\newblock Visualizing deep convolutional neural networks using natural
  pre-images.
\newblock 2016.

\bibitem[McDuff et~al.(2013)McDuff, Kaliouby, Senechal, Amr, Cohn, and
  Picard]{mcduff2013affectiva}
Daniel McDuff, Rana Kaliouby, Thibaud Senechal, May Amr, Jeffrey Cohn, and
  Rosalind Picard.
\newblock Affectiva-mit facial expression dataset (am-fed): Naturalistic and
  spontaneous facial expressions collected.
\newblock In \emph{Proceedings of the IEEE Conference on Computer Vision and
  Pattern Recognition Workshops}, pages 881--888, 2013.

\bibitem[Parkhi et~al.(2015{\natexlab{a}})Parkhi, Vedaldi, and
  Zisserman]{Parkhi15}
O.~M. Parkhi, A.~Vedaldi, and A.~Zisserman.
\newblock Deep face recognition.
\newblock In \emph{British Machine Vision Conference}, 2015{\natexlab{a}}.

\bibitem[Parkhi et~al.(2015{\natexlab{b}})Parkhi, Vedaldi, and
  Zisserman]{parkhi15deep}
O.~M. Parkhi, A.~Vedaldi, and A.~Zisserman.
\newblock Deep face recognition.
\newblock In \emph{British Machine Vision Conference}, 2015{\natexlab{b}}.

\bibitem[Post et~al.(2008)Post, Van~den Assem, Baltussen, and
  Thaler]{post2008deal}
Thierry Post, Martijn~J Van~den Assem, Guido Baltussen, and Richard~H Thaler.
\newblock Deal or no deal? decision making under risk in a large-payoff game
  show.
\newblock \emph{The American economic review}, 98\penalty0 (1):\penalty0
  38--71, 2008.

\bibitem[Rep.~Capuano and Rep.~Jones(Introduced in US House of Representatives,
  02/27/2015)]{CapuanoWAWYA2015}
Michael~E. Rep.~Capuano and Walter B.~{Jr.} Rep.~Jones.
\newblock {We Are Watching You Act}, {H.R.}1164, Introduced in US House of
  Representatives, 02/27/2015.

\bibitem[Russakovsky et~al.(2014)Russakovsky, Deng, Su, Krause, Satheesh, Ma,
  Huang, Karpathy, Khosla, Bernstein, Berg, and Fei-Fei]{russakovsky14imagenet}
Olga Russakovsky, Jia Deng, Hao Su, Jonathan Krause, Sanjeev Satheesh, Sean Ma,
  Zhiheng Huang, Andrej Karpathy, Aditya Khosla, Michael Bernstein,
  Alexander~C. Berg, and Li~Fei-Fei.
\newblock Imagenet large scale visual recognition challenge, 2014.

\bibitem[Schroff et~al.(2015)Schroff, Kalenichenko, and
  Philbin]{schroff2015facenet}
Florian Schroff, Dmitry Kalenichenko, and James Philbin.
\newblock Facenet: A unified embedding for face recognition and clustering.
\newblock In \emph{Proceedings of the IEEE Conference on Computer Vision and
  Pattern Recognition}, pages 815--823, 2015.

\bibitem[Sebe et~al.(2007)Sebe, Lew, Sun, Cohen, Gevers, and
  Huang]{sebe2007authentic}
Nicu Sebe, Michael~S Lew, Yafei Sun, Ira Cohen, Theo Gevers, and Thomas~S
  Huang.
\newblock Authentic facial expression analysis.
\newblock \emph{Image and Vision Computing}, 25\penalty0 (12):\penalty0
  1856--1863, 2007.

\bibitem[Simonyan and Zisserman(2015)]{simonyan15very}
K.~Simonyan and A.~Zisserman.
\newblock Very deep convolutional networks for large-scale image recognition.
\newblock 2015.

\bibitem[Vedaldi and Lenc(2015)]{vedaldi2015matconvnet}
Andrea Vedaldi and Karel Lenc.
\newblock Matconvnet: Convolutional neural networks for matlab.
\newblock In \emph{Proceedings of the 23rd ACM international conference on
  Multimedia}, pages 689--692. ACM, 2015.

\bibitem[Yu and Zhang(2015)]{yu2015image}
Zhiding Yu and Cha Zhang.
\newblock Image based static facial expression recognition with multiple deep
  network learning.
\newblock In \emph{Proceedings of the 2015 ACM on International Conference on
  Multimodal Interaction}, pages 435--442. ACM, 2015.

\bibitem[Zhang and Zhang(2014)]{zhang2014improving}
Cha Zhang and Zhengyou Zhang.
\newblock Improving multiview face detection with multi-task deep convolutional
  neural networks.
\newblock In \emph{IEEE Winter Conference on Applications of Computer Vision},
  pages 1036--1041. IEEE, 2014.

\bibitem[Zhu and Ramanan(2012)]{zhu2012face}
Xiangxin Zhu and Deva Ramanan.
\newblock Face detection, pose estimation, and landmark localization in the
  wild.
\newblock In \emph{Computer Vision and Pattern Recognition (CVPR), 2012 IEEE
  Conference on}, pages 2879--2886. IEEE, 2012.

\end{thebibliography}
\end{document}